\documentclass[letterpaper, 10 pt, conference]{ieeeconf}  

\IEEEoverridecommandlockouts                              

\usepackage{booktabs}
\usepackage{xcolor}
\usepackage{subfigure}
\usepackage{graphicx}
\usepackage{makecell}
\usepackage[
backend=bibtex,
natbib=true,
sorting=none,
useauthor=true,
style=numeric,
maxnames=1,
maxbibnames=25,
minbibnames=25,
]{biblatex}
\DefineBibliographyStrings{english}{andothers = {{et\,al\adddot}},}

\usepackage[
    bookmarks,
    bookmarksopen=true,
    bookmarksnumbered, 
    colorlinks=true,
    linkcolor=black, 
    anchorcolor=black,
    citecolor=black, 
    menucolor=black, 
    urlcolor=black, 
    plainpages=false, 
    pdfpagelabels=true, 
    hypertexnames=true, 
]{hyperref}
\usepackage{cleveref}

\newcommand{\wiedholz}[1]{\textcolor{blue}{}} 

\title{\LARGE \bf
Semantic 3D scene segmentation for robotic assembly process execution
}

\author{Andreas Wiedholz$^{1}$, Stefanie Wucherer$^{1}$ and Simon Dietrich$^{1}$
\thanks{$^{1}$Faculty of Electrical Engineering,
        University of applied Sciences Augsburg, Germany
        {\tt\small <firstname>.<lastname>@hs-augsburg.de}}%
}

\begin{document}

\maketitle
\thispagestyle{empty}
\pagestyle{empty}

\begin{abstract}
Adapting robot programmes to changes in the environment is a well-known industry problem, and it is the reason why many tedious tasks are not automated in small and medium-sized enterprises (SMEs). 
A semantic world model of a robot's previously unknown environment created from point clouds is one way for these companies to automate assembly tasks that are typically performed by humans. 
The semantic segmentation of point clouds for robot manipulators or cobots in industrial environments has received little attention due to a lack of suitable datasets. 
This paper describes a pipeline for creating synthetic point clouds for specific use cases in order to train a model for point cloud semantic segmentation.
We show that models trained with our data achieve high per-class accuracy (\textgreater 90\%) for semantic point cloud segmentation on unseen real-world data.
Our approach is applicable not only to the 3D camera used in training data generation but also to other depth cameras based on different technologies.
The application tested in this work is a industry-related peg-in-the-hole process.
With our approach the necessity of user assistance during a robot's commissioning can be reduced to a minimum.
\end{abstract}

\section{Introduction}
%
Changes in a robot's workspace normally lead to robot programmes needing to be adapted and is often combined with a lot of effort 
This makes the use of a robot inefficient if the robot's task often changes.
This work is part of a research project\footnote[2]{https://www.robominds.de/}\wiedholz{Fußnote behalten?} that aims to simplify building robot programmes that are - based on AI algorithms - robust against environmental changes.
The project develops skills like exploration of a workspace, 6 DoF grasping points and path planning algorithms that deal with physical insecurities.
To test these skills, a real-world demonstrator (see \Cref{fig::demonstrator_hardware}) is built.
In this demonstrator a robot should pick objects from several boxes and place them on a deposition target.
In this use case the deposition target is a board that has the negative of the geometry of each object and is called "shadow board".
This work focuses on exploring the robot's environment and providing path planning and grasping algorithms with the necessary information, e.g. collision areas or target poses.
To do this, collision areas, starting/target robot poses and a process control are provided.

Our approach for the workspace exploration is based on a 3D semantic segmentation of the robot's environment. 
Due to a lack of comparable work and datasets, a semantic scene understanding for robot manipulators, or cobots, in industrial settings is not thoroughly researched \cite{covered_ds}. 
Based on a semantic understanding of the environment, we build an autonomous, task-based process control for the robot.
During an exploration of the robot's workspace, a 3D camera mounted on the robot's flange captures point clouds that are semantically labelled by a neural network.
As there is no suitable dataset available for our use case, the generation of a dataset containing synthetic and real-world data is essential.
This dataset is used to train a model that predicts semantic labels in a point cloud.
Based on this semantic world model, the destination target for grasped objects is exactly located, and positions for the robot in the upcoming assembly task are calculated, so no further teaching of robot poses in the assembly process is required.
The destination target is located using a point cloud registration algorithm and the CAD of the object to be found in the real-world point cloud.
The poses used in the exploration are taught by the user.
Thus, the main contributions of this paper are:
\begin{itemize}
    \item Concept and implementation of a pipeline to generate a dataset containing synthetic point clouds that can be used for the exploration of an unknown workspace of a robot manipulator
    \item Autonomous calculation of robot poses during the exploration using 3D data without intermediate 2D representation to successfully execute an assembly task without the need of human interaction
    \item Validation of the robustness of the model trained with our approach works with the same dataset for point clouds generated by different 3D cameras
\end{itemize}

\section{Related Work}
\subsection{Synthetic data generation}
\label{ssec::synth-data-gen-rel-work}
Since manually creating a real-world dataset is very time-consuming and expensive, the generation of synthetic data is an important task in machine learning applications.
In general, the use of domain randomization has been proven to train neural networks with data in so many variations that the real-world noise only appears as a new unknown variation.
Structured domain randomization was introduced to keep random changes realistic \cite{structured_dr}.
For our approach, semantically labelled point clouds need to be created, and there are several ways to do this.
In general, point clouds can either be generated directly or created based on a RGB-D image.
One implementation for generating semantically labelled point clouds directly is proposed by \cite{blainder}.
Creating a point cloud based on a RGB-D image needs the intrinsic camera parameters and one implementation for this technique is proposed by \cite{denninger2019blenderproc}. 
Both of these implementations are integrated in Blender and are called BlAInder \cite{blainder} and BlenderProc \cite{denninger2019blenderproc}.
With BlAInder, it is possible to simulate (noise-free) Time-of-Flight and LiDAR cameras as they implement their own ray tracing to directly annotate a point in a point cloud with the class of the object that a ray hits first.
The user is limited to the standard Blender depth cameras when using BlenderProc.
NVIDIA IsaacSim\footnote[3]{https://developer.nvidia.com/isaac-sim}, which has high GPU requirements, is another option for simulating 3D cameras and directly generating semantically labelled point clouds. 

Another approach is proposed by \cite{bim2pcl}. 
They use the "Building Information Model" (BIM), which is commonly used in architectural contexts and contains numerous details about texture, material, etc.
They generate highly realistic synthetic data from these models using an automated adaptation and creation of BIM models.

\subsection{Semantic segmentation in robotics}
Most work in robotics with semantic segmentation for scene understanding uses 2D images, and when 3D information is used, it is usually for mobile or service robots rather than robot manipulators \cite{semantic_robotics_survey}.
According to \cite{covered_ds}, the majority of applications that use point cloud semantic segmentation to explore their environment are in the field of mobile robots or autonomous driving because there are corresponding datasets to train their models.
Because of this research gap, they introduced a new dataset that can be used for cobot applications in industrial environments.
Semantic segmentation in manipulation tasks is used more for the bin-picking part of an application, as shown in \cite{pose_estimation_semseg}.
They first take an RGB image, apply semantic segmentation, and project the segmentation results on a point cloud to find graspable objects.

\section{Experimental setup}
\label{ssec::real_world_demonstrator}
\begin{figure}[!ht]
    \centering
    \includegraphics[width=\linewidth]{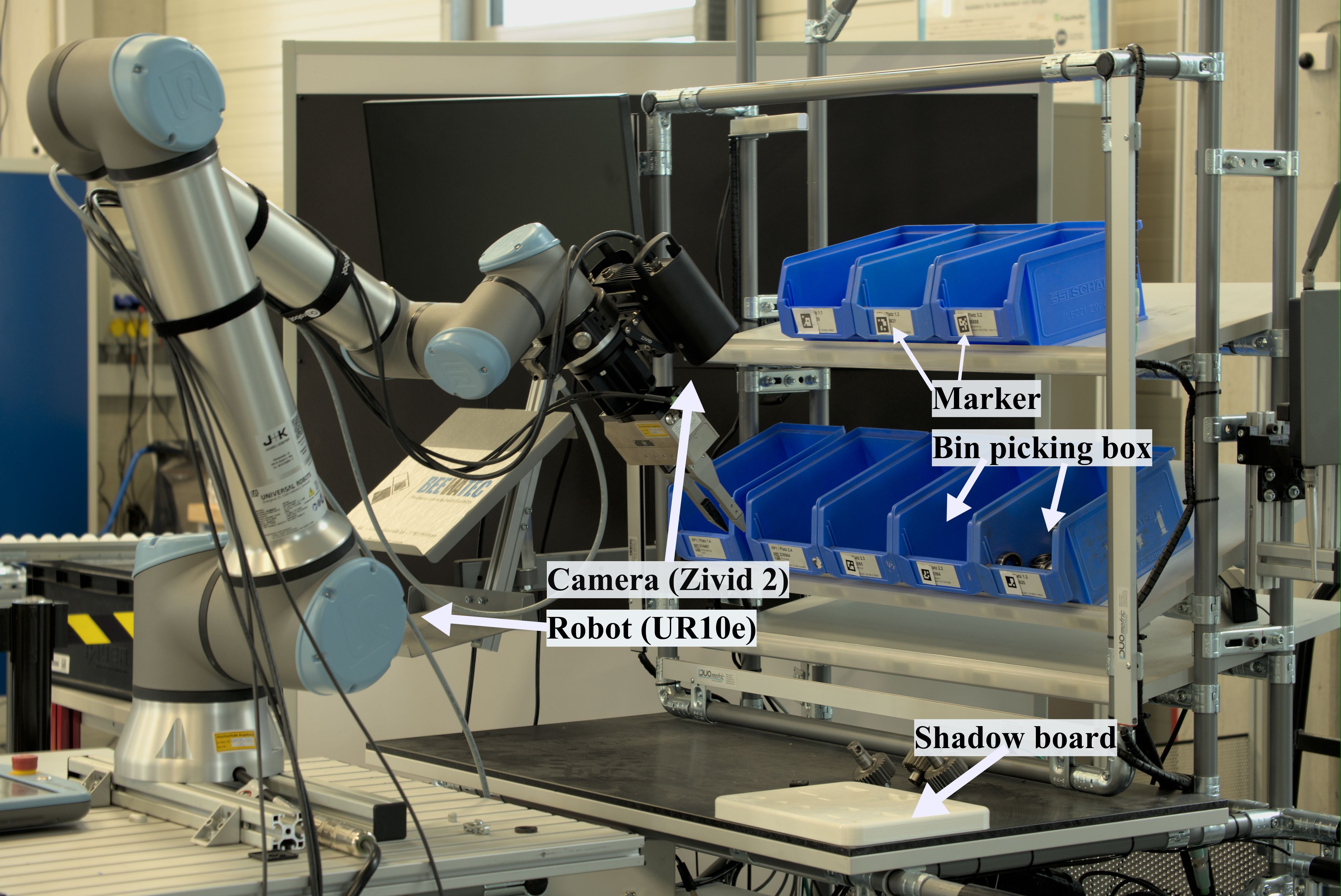}
    \caption{Hardware setup for demonstrator in our use case \wiedholz{Bild mit höherer Auflösung einfügen}}
    \label{fig::demonstrator_hardware}
\end{figure}

To test the workspace exploration, the Pick-and-Place use case from the research project in \Cref{fig::demonstrator_hardware} is used.
The environment is unknown in advance of the robot's commissioning.
Each box is single variety and is labelled with a marker to indicate which object is inside the box.
A high precision 3D camera (Zivid 2) is mounted at the flange of the robot (UR10e) to explore the environment.
We use the Zivid 2 camera as it produces noise-free point clouds and has much higher depth precision than common stereo vision or Time-of-Flight cameras.

\begin{figure}
    \centering
    \subfigure[Process steps for demonstrator]{
    \includegraphics[width=0.9\linewidth]{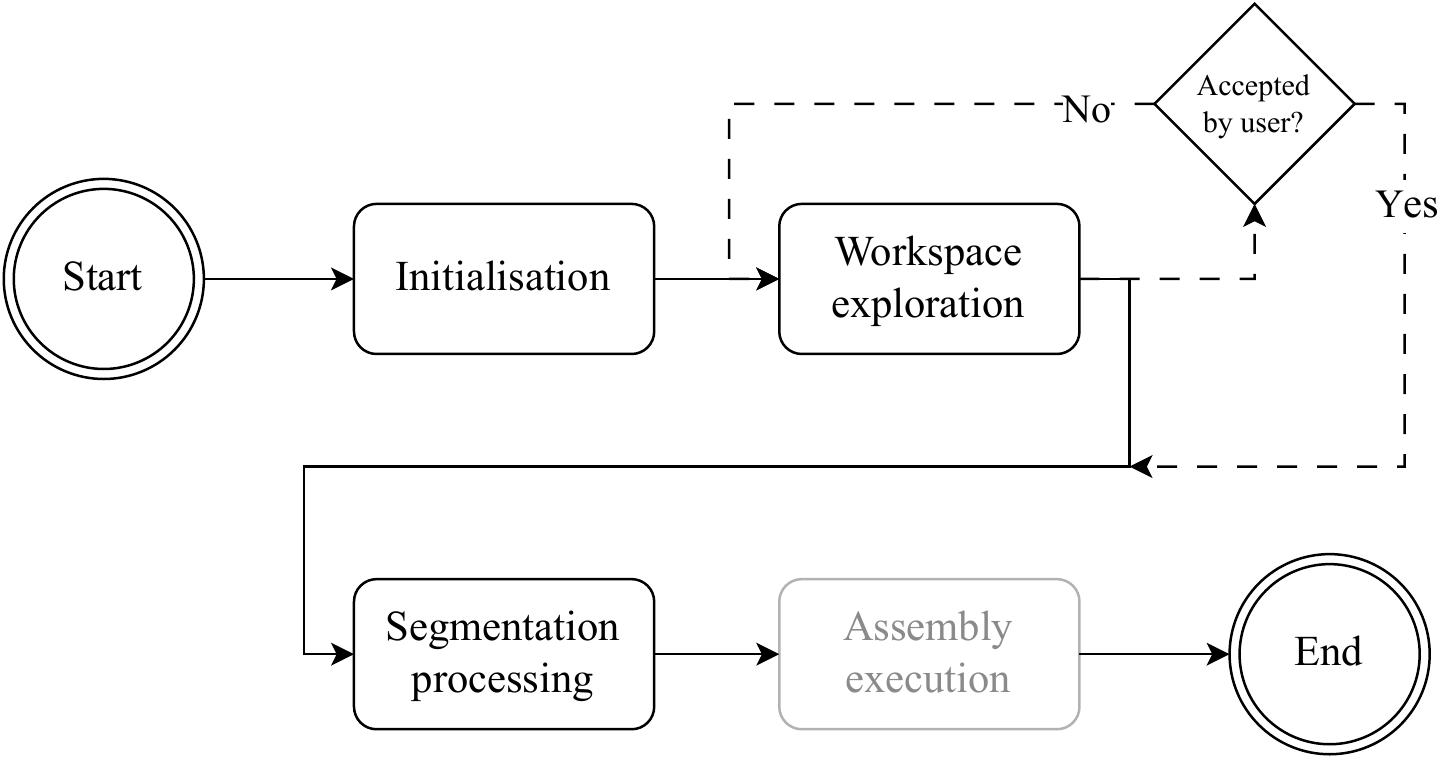}
    \label{pic::process_steps}
    }
    \subfigure[Communication structure]{
    \includegraphics[width=0.9\linewidth]{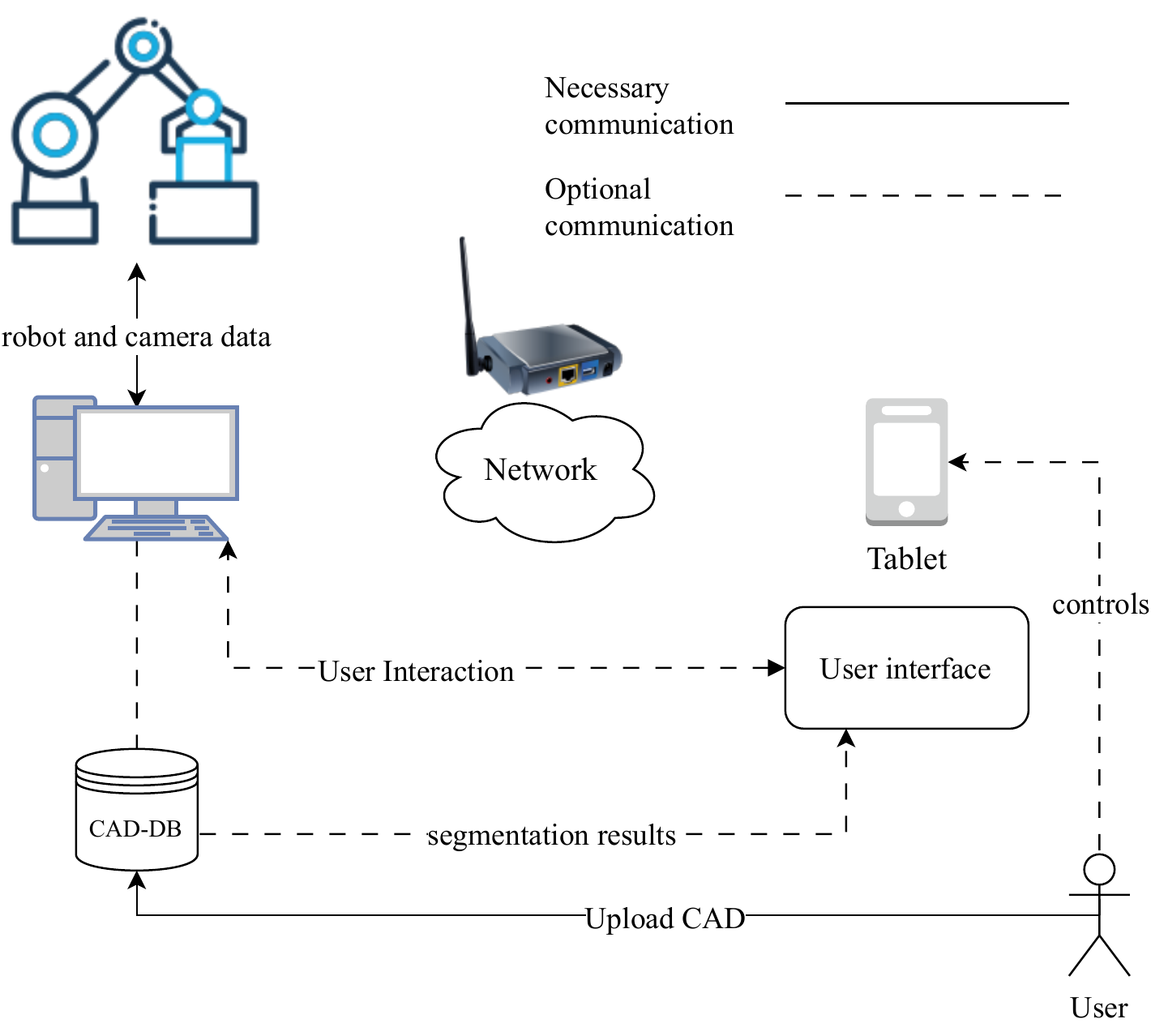}
    \label{pic::communication structure}
    }
    \caption{Hard-/Software structure of the real-world demonstrator}
    \label{fig:overview_components}
\end{figure}

The commission technician moves the robot during initialisation (see \Cref{pic::process_steps}) through the workspace.
While moving the robot, the technician teaches positions that will be used to capture a point cloud. 
To record the working space, several points have to be taught via the robot.
This is done by an operator using the FreeDrive mode on the robot to move it to the corresponding positions and to teach the corresponding points.
The collision areas will be assembled in the robot base frame using octomaps \cite{hornung13auro} created from point clouds captured in the taught poses during workspace exploration step.


\Cref{pic::communication structure} depicts all of the components involved in our setup and how they communicate with one another. 
At the start of the workspace exploration, three docker containerized applications will be launched at the workstation connected to the robot:
\begin{itemize}
    \item process control and calculation of robot poses
    \item semantic segmentation of point clouds
    \item control of the robot via the according driver\footnote[4]{https://github.com/UniversalRobots/Universal\_Robots\_ROS2\_Driver}
\end{itemize}
Every application is ROS 2 integrated as robust inter-process communication can be implemented very easily in this framework.
The training framework proposed by \cite{robert2022dva} - based on \cite{tp3d} - is used for model training and inference for the neural networks.

After the workspace exploration, the user can interact with the process through a user interface that is deployed on a tablet.
Currently, after building a semantic world from all the camera images, the user receives a mesh via a database (see \Cref{pic::communication structure}) reconstructed from the full point cloud and can either reject or accept the segmentation.
If the segmentation result is rejected, the exploration is repeated until the user is satisfied with the result.
The user interaction is implemented to make the system more robust as the results are verified.
For the process itself, user interaction with the system is optional.

\section{Methodology \& implementation}
\subsection{Dataset generation}
\begin{figure*}[t]%
\centering
\subfigure[Bulk box]{%
	\includegraphics[width=0.15\textwidth]{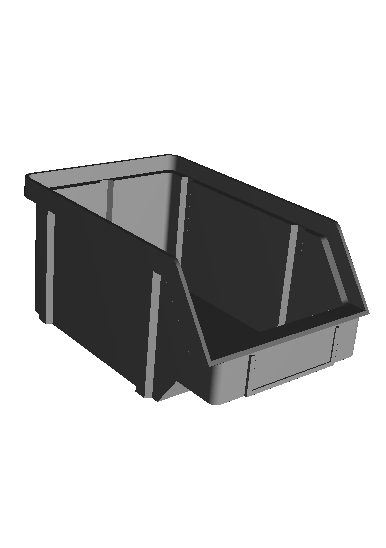}
    \label{fig::bulk-box}
	}%
\subfigure[Storage rack]{%
	\includegraphics[width=0.15\textwidth]{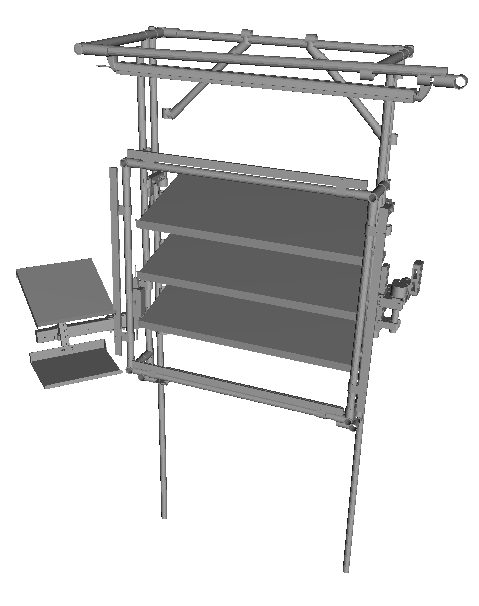}
	}%
\subfigure[Table]{%
	\includegraphics[width=0.15\textwidth]{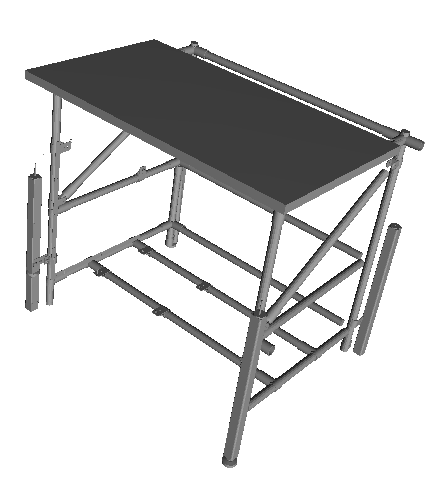}
	}%
\subfigure[Shadow board]{%
	\includegraphics[width=0.15\textwidth]{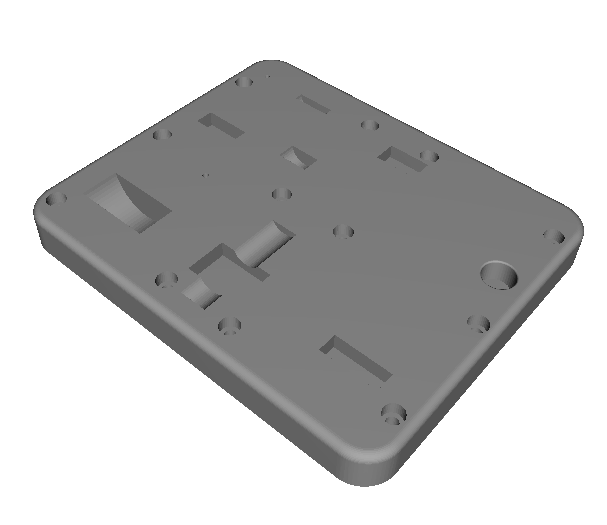}
	\label{pic::shadow_board}
	}%
\subfigure[Synth. point cloud]{%
	\includegraphics[width=0.30\textwidth]{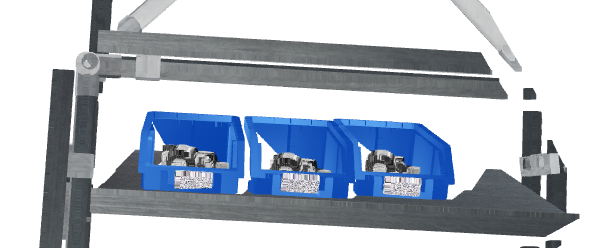}
    \label{pic::synth-point-cloud}
	}%
\caption{Example objects in simulation for datatset generation (\ref{fig::bulk-box} - \ref{pic::shadow_board}) and one example point cloud from the generated dataset (\ref{pic::synth-point-cloud})}
\label{pic::available_ojects}
\end{figure*}

We create a new dataset mostly based on synthetic data.
This dataset contains a total of 2218 semantically labelled point clouds and is split into synthetic data (90\%) and real-world data (10\%).
It contains four classes: storage rack, box/storage bin, shadow board and table (see \Cref{pic::available_ojects}).

\begin{figure}[!ht]
    \centering
    \includegraphics[width=\linewidth]{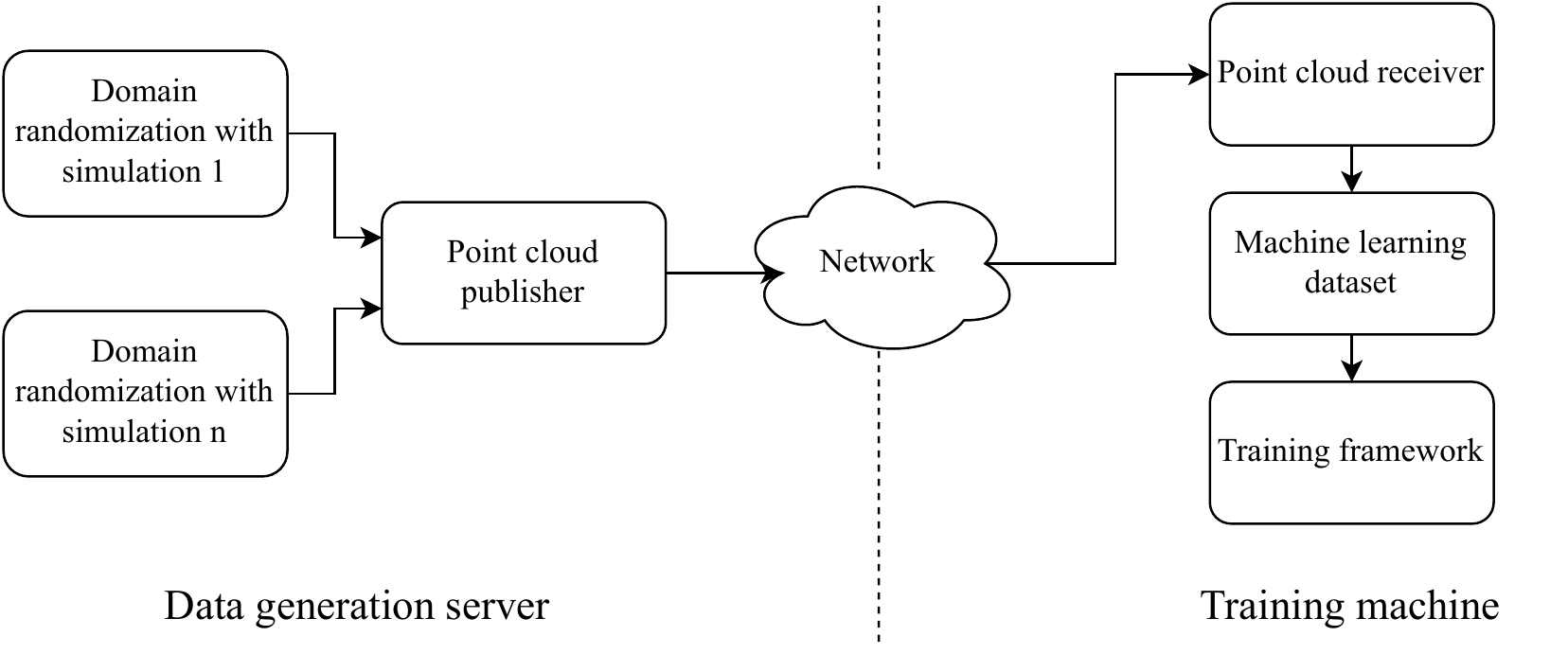}
    \caption{Pipeline from synthetic data generation with BlAInder to train a semantic segmentation model in Torch-Points3D}
    \label{fig::synth-data-generation}
\end{figure}

The pipeline for generating synthetic data created in this work is depicted in \Cref{fig::synth-data-generation}.
We use the Push-Pull pattern, a common pattern in data communication for multiple participants, and combine it with domain randomization in the simulation environment.
Conceptually, this pattern is chosen to simplify the data transport from the data generation application to the training framework.
On one side, the data will be generated based on the idea of structured domain randomization \cite{structured_dr} and a tool for creating automatically labelled point clouds.
The other side receives these data and converts them into a dataset that can be loaded for neural network training.

Blender is chosen as simulation environment and BlAInder \cite{blainder} is chosen to generate semantically labelled point clouds.
For each class in our dataset, several (\textless 5) different 3D models (see \Cref{pic::available_ojects}) are loaded into different Blender scenes to ensure that every combination of 3D models appears in at least one scene.
A label with a marker and metallic gearbox parts are added to some boxes, which will also be classified as boxes because for the exploration itself, we do not need to distinguish between box contents and the box itself.
The marker added to each box encodes information about the objects inside.
It is possible to highly automate the data generation process using these scenes and self-implemented domain randomization methods.
We use BlendTorch's \cite{blendtorch} Push-Pull implementation as it allows to launch multiple instances of Blender at the same time and apply one of our self-implemented domain randomization methods to each scene. 

It is possible to change three kinds of attributes of each object (storage rack, box, shadow board, table) in a scene:
\begin{itemize}
    \item Location, i.e. rotating or translating the object
    \item Geometric properties, i.e. scaling the object
    \item Visual properties, i.e. changing material attributes (colour, specularity, transparency, roughness) or the image texture applied to the surface
\end{itemize}

As the camera position in the real-world application is determined by the user, it is impossible to predict from what point of view the camera will capture the environment during data generation.
As a result, the simulation must account for every possible camera position and orientation. 

The training machine (see \Cref{fig::synth-data-generation}) receives the semantically labelled point clouds and stores them in a PyTorch dataset.
This dataset is integrated in TorchPoints3D \cite{tp3d} - a framework for training several neural network architectures on different point cloud tasks - to compare several network architectures on our data.
In \Cref{sec::results} the evaluation of these models on real-world datasets is described.

Real-world point clouds added to the dataset were captured with the Zivid 2 camera and manually labelled\footnote[5]{https://segments.ai/}.
To increase the number of real-world point clouds without having to manually label many more new captures, data augmentation methods are used.
Therefore, we introduce noise in RGB and XYZ space, rotate, and translate the point clouds.
For our use case, 40 real camera point clouds have empirically proven to be sufficient as shown in \Cref{sec::results}.

\subsection{Assembly process}
For the assembly process we need the following data:
\begin{itemize}
    \item 3D camera and transformation matrix from camera to robot flange frame and the current robot flange to robot base frame
    \item CAD objects of all objects used in assembly process (grasped objects, deposition object)
\end{itemize}

The 3D camera is needed to explore the workspace and locate the shadow board in the environment. 
Using the CAD data of the gripped object and the shadow board - in this case, the shadow board - it is possible to calculate a target position. 
From this target position, the robot is able to successfully place the gripped object.


We distinguish between dynamic and static computations.
Anything other than finding the target position to place a given object in the shadow board relative to the shadow board's coordinate frame is considered dynamic computation.
However, as long as the environment does not change, i.e. the position of boxes, collision areas, etc. remain constant, the result of the workspace exploration can be reused and is static. 
A database is set up for reusable results, e.g. target position for depositing a specific object (see \Cref{pic::communication structure}).
Each graspable object with its marker ID is stored in this database, and after identifying the marker of a box, the system knows the target position, which saves time during the assembly process.

We define three main steps in our approach that are required for a complete understanding of the scene and detailed knowledge of the environment (see \Cref{pic::process_steps}). 
\subsubsection{Initialisation}
A human moves the robot to certain positions from which the camera is supposed to take a capture. 
The robot is put into free-drive mode, and the user can store robot positions for camera captures by pressing buttons added to the flange.
\subsubsection{Workspace exploration}
The user-taught positions are used in the workspace exploration step when the robot autonomously moves to each position and takes a capture.
Each capture is processed individually and fed into the semantic segmentation network.
We first downsample the point cloud to a voxel size of one millimetre and apply the Statistical Outlier Removal (SOR) filter first introduced by the Point Cloud Library \cite{pcl_lib}.
Then we randomly sample 8192 points and get the label predictions from the semantic segmentation network.
We choose to use 8192 points as it has empirically been proven to be a sufficient number of points to get high accuracy of correct predictions.
In post-processing, we use the predictions to generate labels for the points that were left out.
We annotate the unlabelled points as the algorithms in the segmentation processing (see \Cref{ssec::seg_processing}) needs more than 8192 points to work reliably.
The labelled point cloud is then transformed into the robot's base frame. 
\subsubsection{Segmentation processing}
\label{ssec::seg_processing}
This step involves performing additional computations based on the previously constructed semantic world.
The segmentation processing step is used to generate robot poses for assembly execution.

First, all points classified as boxes are selected and the individual instances of the boxes are determined. 
Since we assume that the label with the marker is white and the box has a different colour, we select all the white points and apply the DBSCAN algorithm \cite{dbscan}. 
Thus, we can find single instances of the label to indicate an instance of a box, generate robot poses for e.g. bin picking and store it for quick access during assembly.

The next step is to select all the points that have been classified as shadow board.
The robot returns to this position and takes another capture.
This point cloud can be used to register a CAD model of the shadow board in order to determine the exact position and orientation of the shadow board.
To use registration algorithms, a point cloud is uniformly sampled from the CAD of the shadow board.
For this registration task, the feature-matching based on RANSAC implemented in Open3D \cite{open3d} is used as it performs more reliably than Iterative Closest Point (ICP) or Fast Global Registration (FGR) in this task.
The features to be matched are FPFH (Fast Point Feature Histograms) \cite{fpfh} features computed in advance for each point in the two point clouds.

The FPFH feature matching algorithm can be applied not only to find the shadow board in a point cloud but also to find objects in the shadow board and generate the target position for grasped objects in the shadow board.
To do this, a CAD model is created with all objects in the correct position and two point clouds are sampled from the corresponding CAD models.
These two point clouds can be used to fit the grasped object into the shadow board.
In this case, sampling 100,000 points from the shadow board and sampling 5000 points from the object to be fitted into the shadow board worked reliably.


\subsubsection{Assembly execution}
In this step, objects are grasped and deposited in the shadow board.
As the grasp planning and the placement of the grasped object in the shadow board are challenges that will be investigated in future work.

\section{Results and Validation}
\label{sec::results}
\subsection{Workspace exploration}
We conduct two experiments to validate the workspace exploration.
First, we investigate the effect of adding real-world data to the synthetic dataset.
The models we train for this experiment are PointNet \cite{pointnet}, PPNet \cite{ppnet}, PVCNN \cite{pvcnn} and KPConv \cite{Thomas_2019_ICCV} as these are common network architectures achieving sufficient accuracies on S3DIS benchmark and are integrated in the TorchPoints3D framework.
The best model from this evaluation is used to investigate how well it performs on other 3D camera systems that are cheaper than the currently used Zivid 2.

\begin{table}[htbp]
\centering
\caption{Mean Accuracy and mean Intersection over Union for different models on a real-world test dataset}
\label{tab::macc_miou_diff_ds}
\begin{tabular}{@{}lcc|cc@{}}
\toprule
\multicolumn{1}{c}{\textbf{Model}} & \multicolumn{2}{c|}{\textbf{mAcc}} & \multicolumn{2}{c}{\textbf{mIoU}} \\ \midrule
\textbf{}                       & \textbf{Real \& Synth} & \textbf{Synth} & \textbf{Real \& Synth} & \textbf{Synth} \\ \midrule
KPConv  \cite{Thomas_2019_ICCV} & 97.9         & \textbf{73.59}          & 96.28        & \textbf{58.24}          \\
PointNet \cite{pointnet}        & 91.95        & 64.39          & 86.96        & 47.61          \\
PVCNN \cite{pvcnn}              & \textbf{97.91}        & 60.73          & \textbf{96.31}        & 43.32          \\
PPNet \cite{ppnet}              & 96.25        & 62.24          & 95.9         & 45.6           \\ \bottomrule
\end{tabular}
\end{table}

\Cref{tab::macc_miou_diff_ds} shows the mean per class accuracy and mean Intersection over Union on a test dataset containing only real-world data (8192 points sampled per point cloud).
Four different models were each trained on two different training datasets.
"Real \& Synth" means that the training dataset also contains real-world data and "Synth" means that only synthetic data were used.
In general, models perform much better when being trained on a dataset that also contains real-world data, which indicates that there is a high Sim2Real gap.
As the results with very few real-world point clouds added to the training set are sufficient for our use case, we did not further investigate this gap. 
Since there is no noticeable difference between KPConv and PVCNN on data generated by the Zivid 2 camera, we use both models when investigating the performance on point clouds captured by other camera systems. 

The performance of PVCNN and KPConv on point clouds generated by five camera systems is shown in \Cref{tab:macc_different_cameras}.
As validation cameras, three stereo vision cameras (Intel Realsense D415, D435i and D455), one LiDAR camera (Intel Realsense L515) and one Time-of-Flight camera (Microsoft Azure Kinect) are used.
The results show that both networks achieve comparable results for low-budget cameras as they do for the Zivid 2 camera, despite KPConv having slightly better accuracy for all cameras.
This shows that for a workspace exploration, independently from the camera, our approach leads to an accurate and robust process.

\begin{table}[htbp]
    \centering
    \caption{Accuracy of PVCNN \cite{pvcnn} and KPConv \cite{Thomas_2019_ICCV} on data generated by different 3D cameras}
    \label{tab:macc_different_cameras}
    \begin{tabular}{@{}cc|cccc@{}}
    \toprule
    \textbf{Camera} & \textbf{mAcc} & \makecell{\makecell{\textbf{Storage}}\\ \textbf{rack}} & \textbf{KLT} & \textbf{Shadow board} & \textbf{Table} \\ \midrule
    \noalign{\smallskip}
    \multicolumn{6}{c}{PVCNN} \\ \midrule
RS L515      & 91.02 & 89.95    & 88.65     & 89.97     & 95.54          \\
Azure Kinect & 94.74 & 91.27    & 95.98     & 94.54     & 97.17          \\
RS D415      & 95.45 & 90.76    & 97.38     & 94.16     & 99.49          \\
RS D455      & 95.81 & 92.55    & 94.2      & 97.1      & 99.41          \\
RS D435i     & 92.54 & 91.29    & 90.83     & 88.29     & \textbf{99.78}          \\ 
Zivid 2      & 97.16 & 96.75    & 96.92     & \textbf{98.68}     & 96.30         \\ \midrule

    \noalign{\smallskip}
    \multicolumn{6}{c}{KPConv} \\ \midrule
RS L515      & 94.18 & 95.58    & 94.41     & 91.64     & 95.07 \\
Azure Kinect & 94.99 & 92.88    & 96.88     & 93.34     & 96.86 \\ 
RS D415      & 97.65 & 97.32    & \textbf{98.54}     & 96.05     & 98.68 \\
RS D455      & 97.08 & 95.79    & 94.9      & 98.07     & 99.55 \\
RS D435i     & 94.15 & 90.73    & 93.17     & 97.19     & 95.49 \\
Zivid 2      & \textbf{97.89} & \textbf{98.59}    & 97.74     & 98.40     & 96.83 \\ \bottomrule
    \end{tabular}
\end{table}

\subsection{Segmentation processing}
\begin{figure}[htb]%
    \centering
    \subfigure[MS Azure Kinect, fitness 0.39]{%
        \label{pic::azure_fitness_0.39}
        \includegraphics[width=0.4\linewidth]{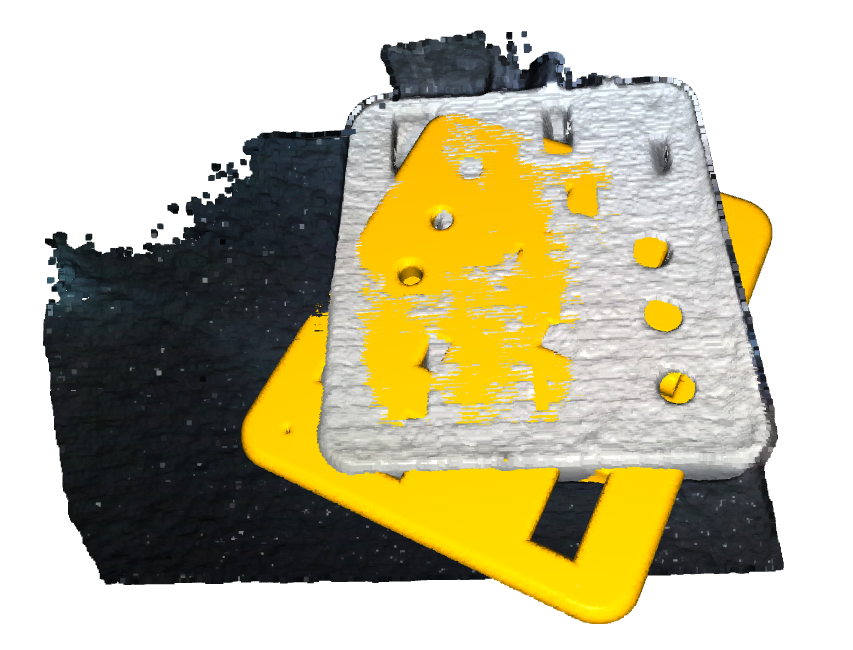}
        }%
    \hfill
    \subfigure[Zivid 2, fitness 0.84]{%
        \label{pic::zivid_fitness_0.84}
        \includegraphics[width=0.4\linewidth]{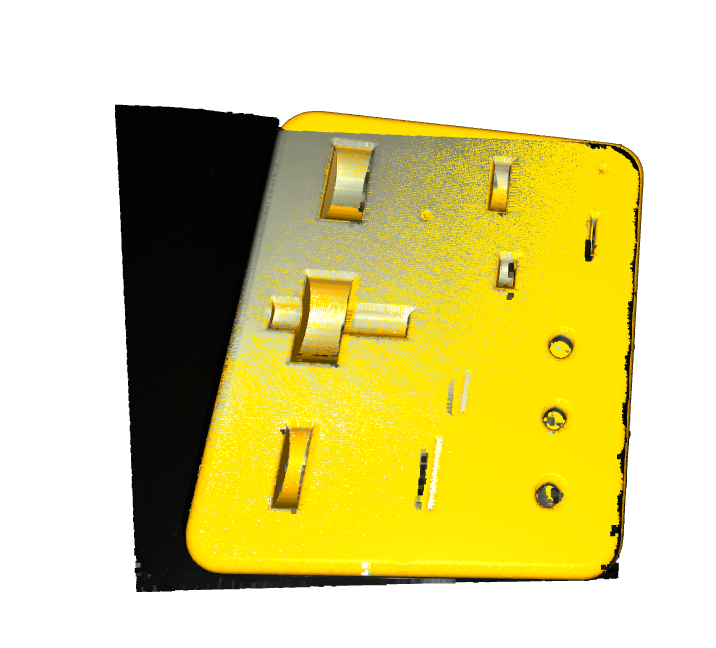}
        }%
    \caption{Shadow board registration for different 3d camera systems}
    \label{pic::sb_registration}
\end{figure}

The second part of the validation also refers to the comparison of the Zivid 2 camera to other cameras.
Finding the shadow board in the real world is a crucial part of this work, as a translation error of only a few millimetres or a rotation error of a few degrees can lead to failure in the subsequent assembly process.
The metric used here is fitness, which is defined by the percentage of points from the CAD model that were found in the real-world point cloud, i.e. the valid range is [0...1].
\Cref{tab:fitness_different_cameras} displays the fitness of the registration algorithm for all investigated camera systems.
We captured 10 point clouds from different angles of the shadow board and tried 8 times to match the CAD model in this point cloud.
Every point cloud was downsampled with a voxel size of 1mm.
\Cref{pic::sb_registration} shows that a fitness level of 0.39 is not enough to fit the CAD model sufficiently into a point cloud.
Based on empirical experiments, we found 0.75 to be a suitable threshold to trust a fit in the application.
As can be seen in \Cref{tab:fitness_different_cameras}, that means that none of the low-budget cameras can be used to fit a CAD model reliably in a point cloud. 
\Cref{tab:fitness_different_cameras} also shows that this algorithm works better if the point cloud contains more points.
For this task - when using a low-budget camera - it is necessary to find a more robust registration algorithm.

\begin{table}[htbp]
\centering
\caption{Fitness for shadow board registration on different 3d cameras (RS = Intel Realsense, \# points refers to the number of points in the real point cloud)}
\label{tab:fitness_different_cameras}
\begin{tabular}{@{}ccc@{}}
\toprule
\textbf{Camera} & \textbf{Fitness} & \textbf{Technology}\\ \midrule
RS D455         & $0.03 \pm 0.13$  & Stereo vision      \\
RS D435i        & $0.06 \pm 0.15$  & Stereo vision      \\
RS D415         & $0.02 \pm 0.1$   & Stereo vision      \\ 
Azure Kinect    & $0.31 \pm 0.21$  & Time-of-Flight     \\
RS L515         & $0.47 \pm 0.13$  & LiDAR              \\
Zivid 2         & $0.88 \pm 0.11$  & Structured light   \\ \bottomrule
\end{tabular}
\end{table}

\section{Conclusion \& outlook}
This paper proposed a method for semantically exploring a robot's workspace in an industrial environment while requiring minimal user interaction.
A dataset containing both synthetic and real-world data is necessary to train two models capable of semantically identifying objects in a point cloud. 
Only 40 manually labelled point clouds are required to achieve very high per-class accuracy on previously unseen real-world data.
The networks are trained from scratch and not only perform for the camera that has been used for generating training data but also for other 3D cameras that rely on different technologies.
Thanks to the proposed data generation pipeline, a new use case only requires the creation of new scenes in Blender and a few real-world captures to train a model.

For the semantic segmentation, the strengths of the Zivid 2 camera are not used but in the registration task the high quality point cloud is necessary.
Every required robot pose for an assembly process in an unknown static environment can be precalculated using the semantic world built with point clouds.

In future work, more robust algorithms for registering a CAD model in a real-world point cloud should be investigated in order to reduce reliance on a large number of points and a high-quality point cloud.
To reduce user interaction, an autonomous, frontier-based exploration strategy should be implemented.
Furthermore, this approach should be transferred and tested in different scenarios. 

%
\section*{Acknowledgment}
This research was funded by the Bavarian Ministry of Economic Affairs, Regional Development and Energy under the research program "Information and Communication technology".
\printbibliography

\end{document}